\documentclass[lettersize,journal]{IEEEtran}
\usepackage{amsmath,amsfonts}
\ifCLASSOPTIONcompsoc
\usepackage[caption=false,font=normalsize,labelfont=sf,textfont=sf]{subfig}
\usepackage{pifont}
\else
 \usepackage[caption=false,font=footnotesize]{subfig}
 \usepackage{caption}
\fi
\usepackage{url}

\usepackage{booktabs}
\usepackage{adjustbox}
\usepackage{multirow, multicol}
\usepackage[table,xcdraw]{xcolor}
\usepackage{pifont}

\begin{document}

%
\title{In-context Learning of Vision Language Models for Detection of Physical and Digital Attacks against Face Recognition Systems}
%
%
%

\author{Lazaro Janier Gonzalez-Soler,
        Maciej Salwowski,
        and~Christoph Busch,~\IEEEmembership{Senior Member,~IEEE}
\thanks{Lazaro Janier Gonzalez-Soler and Christoph Busch are with da/sec - Biometrics and Security Research Group, Darmstadt, Germany e-mail: (\{lazaro-janier.gonzalez-soler,christoph.busch\}@h-da.de).}
\thanks{Maciej Salwowski was with Technical University of Denmark, Denmark e-mail: (s223525@student.dtu.dk).}
}

\markboth{}%
{Shell \MakeLowercase{\textit{et al.}}: A Sample Article Using IEEEtran.cls for IEEE Journals}


\maketitle

\newcommand{\xmark}{\text{\ding{55}}}

\begin{abstract}
Recent advances in biometric systems have significantly improved the detection and prevention of fraudulent activities. However, as detection methods improve, attack techniques become increasingly sophisticated. Attacks on face recognition systems can be broadly divided into physical and digital approaches. Traditionally, deep learning models have been the primary defence against such attacks. While these models perform exceptionally well in scenarios for which they have been trained, they often struggle to adapt to different types of attacks or varying environmental conditions. These subsystems require substantial amounts of training data to achieve reliable performance, yet biometric data collection faces significant challenges, including privacy concerns and the logistical difficulties of capturing diverse attack scenarios under controlled conditions. This work investigates the application of Vision Language Models (VLM) and proposes an in-context learning framework for detecting physical presentation attacks and digital morphing attacks in biometric systems. Focusing on open-source models, the first systematic framework for the quantitative evaluation of VLMs in security-critical scenarios through in-context learning techniques is established. The experimental evaluation conducted on freely available databases demonstrates that the proposed subsystem achieves competitive performance for physical and digital attack detection, outperforming some of the traditional CNNs without resource-intensive training. The experimental results validate the proposed framework as a promising tool for improving generalisation in attack detection.
\end{abstract}


%

\section{Introduction}

\IEEEPARstart{F}{acial} recognition has become one of the most common methods of identification of individuals in modern society, with authentication processes playing a crucial role in numerous daily activities. The critical nature of these processes, especially in security and access control, drives researchers to improve the reliability and security of such systems continuously.

Recent progress in biometric systems has led to substantial improvements in the accuracy and robustness of fraud detection mechanisms. Nevertheless, the continual evolution of these systems is paralleled by increasingly sophisticated and adaptive attack strategies. According to the International Standard ISO/IEC 30107-3 for biometric presentation attack detection (PAD)~\cite{ISO-IEC-30107-3-PAD-metrics-2023}, nine different attack points can interfere with the normal operation of facial recognition systems. Two categories of these attacks are broadly defined in physical and digital approaches. In the physical realm, attack presentations are the most prevalent, utilising various presentation attack instrument species (PAIs) such as a printed face image, a video replay or other impersonating artefacts to deceive recognition systems. Digital attacks~\footnote{Strictly speaking a morphing attack is a digital manipulation and a subsequent printing/scaning process that completes the attack vector as physical attack instrument.} have become equally worrying, with advanced tools facilitating the creation of very convincing morph images that can be used simultaneously by two users to deceive recognition systems as if they were a single subject.

Traditionally, deep learning models have been the primary defence against such attacks. While these models perform exceptionally well in scenarios for which they have been trained, they often struggle to adapt to different attack types (i.e., unknown PAI species in the context of presentation attack detection (PAD) or unknown morphing tools in the case of morphing attack detection (MAD) or varying environmental conditions~\cite{GonzalezSoler-FVFacePAD-IET-2021}. Current attack detection systems require substantial amounts of training data to achieve reliable performance, yet biometric data collection faces significant challenges, including privacy concerns and the logistical difficulties of capturing diverse attack scenarios under controlled conditions \cite{Caldeira-MADtion-WACV-2025,Ozgur-FoundPAD-ArXiv-2025}. This limitation poses a significant challenge in real-world applications, especially when the nature of an attack is unknown. The process of developing specialised models is time-consuming, expensive, and requires extensive data collection.

To overcome the above limitations, the emergence of vision language models (VLM) offers a promising alternative. These models, trained on vast datasets, can handle complex questions and adapt to various scenarios~\cite{Zhang-VLMsSurvey-ArXiv-2024}, proving a complementary detection approach, which relies solely on the VLM expertise. VLMs thus offer potential to address the challenges of generalisation in biometric attack detection. While preliminary research has explored VLMs for PAD~\cite{Shi-SHIELD-ArXiv-2023,Komaty-ChatGPTPAD-ArXiv-2025} through qualitative assessments, these studies were lacking rigorous quantitative assessment using standardised metrics and were limited to specific attack scenarios. In the case of MAD, very limited research has focused solely on the capabilities of VLMs for single morphing attack detection (S-MAD)~\cite{Zhang-ChatGPTSMAD-2025}.

This paper fills critical gaps in the literature by performing the first comprehensive quantitative analysis of VLMs for both physical (presentation) and digital (morphing) attack detection. By adapting models with contextual knowledge injected only during inference, an in-context learning framework is proposed. The main contributions of our work are:

\begin{itemize}
    \item In-context learning conceptual approaches for physical and digital attack detection (i.e., PAD and S-MAD), which are capable of detecting attack presentations and morphing attacks, respectively, without the need for training: only up to 9 samples are used during network inference.

    \item An extensive review of the state‐of‐the‐art techniques employed for facial PAD and MAD. We mainly emphasise those methods focused on facial PAD and MAD generalisation
    
    \item In-depth analysis of learning performance in the context of VLMs consisting of less than 8 billion parameters for PAD and S-MAD. Contrary to current studies on VLMs~\cite{Zhang-ChatGPTSMAD-2025,Komaty-ChatGPTPAD-ArXiv-2025}, which are based on a maximum of two inference learning shots, in our work, we evaluate up to 9 inference shots.

    \item Extensive evaluation in compliance with metrics defined in the International Standards ISO/IEC 30107-3~\cite{ISO-IEC-30107-3-PAD-metrics-2023} for biometric PAD and ISO/IEC 20059~\cite{ISO-IEC-20059} for MAD of the proposed approaches for different cross-database scenarios. Experimental evaluations show that the proposed framework can achieve state-of-the-art performance in different protocols and outperform baselines.
\end{itemize}

 The remainder of this paper is structured as follows: Related work is summarised in Sect.~\ref{sec:related_work}. In Sect.~\ref{sec:framework}, the conceptual framework based on in-context learning of VLMs for PAD and MAD is described. The experimental setup is summarised in Sect.~\ref{sec:exp_setup}. Experimental results, including the foundation model assessment, as well as a benchmark of the proposed PAD framework on challenging settings, are presented in Sect~\ref{sec:results}. Conclusion and future work directions are summarised in Sect.~\ref{sec:conclusions}.

\section{Related Work}
\label{sec:related_work}

Unlike traditional authentication systems, biometric systems eliminate the need for individuals to remember passwords or carry physical tokens like ID cards or tags. While this reduces the risk of repudiation disputes, biometric systems can still be vulnerable to various forms of manipulation and deception~\cite{ISO-IEC-30107-3-PAD-metrics-2023}. According to \cite{ISO-IEC-30107-3-PAD-metrics-2023}, biometric systems are vulnerable to attacks at nine critical points, which can be broadly categorised into direct and indirect attacks. Direct attacks refer only to sensor attacks (e.g., attack presentations) that do not require any expert knowledge and involve presenting fake biometric traits (e.g., synthetic fingerprints or facial images). In contrast, indirect attacks target the system's internal components and require knowledge of its operation. These include intercepting communication channels to replay or tamper with biometric data (e.g., using morphing images), compromising feature extraction and comparison subsystems. While direct attacks exploit the physical vulnerability of the capture device, indirect attacks challenge the digital and logical security of the system, making comprehensive protection a critical aspect of biometric system design.

\subsection{Presentation Attack Detection}

Attacks that require minimal technical knowledge of the system are known as attack presentations (AP) and typically target the sensor, exploiting its vulnerabilities through simple methods such as the use of a printed facial mask or a synthetic fingerprint. The fabrication of APs includes tools or materials designed to replicate or imitate a legitimate user’s facial traits. Common examples are photographs, video replays, 3D-printed masks, or silicone masks. A malicious subject might use a high-quality printout of a facial image or a carefully crafted mask to easily bypass the system’s authentication process.

To mitigate said threats, the former PAD approaches relied on the analysis of handcrafted features, which detected, among other aspects, texture inconsistencies between PAIs and bona fide presentations (BP). However, with the introduction and success of deep learning in many computer vision and pattern recognition tasks, new PAD subsystems evolved from those primary feature analysis~\cite{Arashloo-PAD-BSIFfusion-2015,Gonzalez-PAD-FVencForFacePAD-BIOSIG-2020,GonzalezSoler-UnkownAttacksFace-IET-Biometrics-2021,Raghavendra-PAD-MS-LPQ-2018} to the development of powerful convolutional neural networks (CNNs)~\cite{Fang-PatchWise-IJCB-2022,Fang-CF-PAD-WCACV-2024,George-PAD-DeepPixelBis-ICB-2019}, and vision transformers~\cite{George-VitEffPAD-IJCB-2021,Ozgur-FoundPAD-ArXiv-2025,GonzalezSoler_ZeroFoundPAD_FG_2025}. 

Back in 2014, Yang \textit{et al.}~\cite{Yang-PAD-CNNApplicability-ArXiv-2014} finetuned ImageNet pretrained CaffeNet~\cite{Jia-PAD-CaffeNet-2014} and VGG-face~\cite{Parkhi-VGG-Face-2015} models for PAD. Following this idea, Xu~\textit{et al.}~\cite{Xu-PAD-LSTM-2015} combined Long Short-Term Memory (LSTM) units with CNNs to learn temporal features from face videos. Sanghvi \textit{et al.}~\cite{Sanghvi-MixNet-ICPR-2021} improved generalisability to unseen attacks by combining three CNN sub-architectures, one for each common PAI species, i.e. print, replay and mask attacks. Fang \textit{et al.}~\cite{Fang-LMFD-WCACV-2022} proposed a hierarchical attention module integration to merge information from two streams at different stages, considering the nature of deep features in different layers of the CNN. Some techniques~\cite{Chen-DualStream-3dMask-CVPR-2021,Liu-ContextAware-TIFS-2022} have also proposed CNNs to analyse properties in 3D mask attacks based on the fact that 2D face PAD algorithms suffer from a significant degradation of detection performance in this type of PAI species. Since acquisition properties such as facial appearance, pose, lighting, capture devices, PAI species and even subjects vary between datasets, several major facial PAD approaches have recently explored domain adaptation (DA) to align features from two different domains~\cite{Fang-CF-PAD-WCACV-2024,Li-KnownledgeDistillation-TIFS-2022,deFreitas-FaceDomainAdaptation-2019,Wang-PAD-AdvDomainAdaptation-TIFS-2020,Wang-MDIL-AI-2024,Yang-PAD-PersonSpecDA-TIFS-2015}.

\subsection{Morphing Attack Detection}

\begin{table}[!t]
	\centering
	\caption{A summary of databases used in our experiments.}
	\label{tab:DB}
    \begin{adjustbox}{width=\linewidth}
	\begin{tabular}{r c r c c l} \toprule \toprule
    \multicolumn{6}{c}{\textbf{PAD}} \\ 
            \textbf{DB}     &   \textbf{\#Videos}     &    \textbf{Split}    &  \textbf{\#BP} &  \textbf{\#AP}  &  \textbf{PAI species}  \\
\midrule
\midrule
\multirow{2}{*}{CASIA-FASD~\cite{Zhang-CASIAFASD-ICB-2012}} & \multirow{2}{*}{600} 	  &     Train     &     60      &  180       &	\multirow{2}{*}{\shortstack[l]{Warped photo (Printed attack), \\ Cut photo, Video replay}}   \\ 
	                	& 						  &	    Test      &     90      &  270       &     \\ 
\midrule
\multirow{3}{*}{REPLAY-ATTACK~\cite{Chingovska-REPLAYATTACK-BIOSIG-2012}} & \multirow{3}{*}{1,200}          &     Train  &     60      &  300    & \multirow{3}{*}{\shortstack[l]{Printed attacks, Photo replay, \\ Video replay}}  \\
                               &			           &     Dev    &     60      &  300    &          \\ 
                               &			           &     Test   &     80      &  400    &         \\ 
\midrule
\multirow{3}{*}{OULU-NPU~\cite{Boulkenafet-OULUNPU-FG-2017}} & \multirow{3}{*}{4,950}    &     Train  &     360     &  1,440  & \multirow{3}{*}{\shortstack[l]{Printed attacks, Video replay}}  \\
	                   &                           &     Dev    &     270     &  1,080  &                     \\
                          &                           &     Test   &     360     &  1,440  &                   \\
\midrule
\multirow{2}{*}{MSU-FASD~\cite{Wen-MSUFASD-TIFS-2015}} 	& \multirow{2}{*}{440} 	  &     Train  &     30      &  90     & \multirow{2}{*}{\shortstack[l]{Printed attacks, \\ Video replay}}  \\
						  &					        &     Test   &     40      &  120    &                 \\
\midrule
\midrule
\multicolumn{6}{c}{\textbf{MAD}} \\ 
            \textbf{DB}     &   \textbf{\#Images}     &    \textbf{Split}    &  \textbf{\#BP} &  \textbf{\#MA}  &  \textbf{Morphing tools}    \\
\midrule
\midrule
\multirow{2}{*}{FERET~\cite{Phillips-FERET-1998}} & \multirow{2}{*}{3437} 	  &     \cellcolor{gray!15}     &     \multirow{2}{*}{1,321}      &  \multirow{2}{*}{2,116}       &	\multirow{4}{*}{\shortstack[l]{FaceFusion, UBO, \\ FaceMorpher, \\ and OpenCV}} \\ 
	                  & 						&	  \cellcolor{gray!15}      &           &         &      \\ 
\cmidrule{1-5}
\multirow{2}{*}{FRGCv2~\cite{Phillips-FRGC-2005}} & \multirow{2}{*}{6,566} 	 &      \cellcolor{gray!15}     &   \multirow{2}{*}{2,710} &  \multirow{2}{*}{3,856} &	 \\ 
	               & 						  &	    \cellcolor{gray!15}      &                        &         &          \\ 
\bottomrule \bottomrule
	 \end{tabular}
     \end{adjustbox}
\end{table}


Morphing attacks (MA) have been identified as one of the most critical threats to biometric systems by the National Institute of Standards and Technology (NIST)~\cite{Ngan-MAD-NIST-2025}, as they exploit the vulnerabilities in biometric enrolment and verification processes, particularly in systems that rely on facial recognition. By creating a morphed image that blends the features of multiple individuals, attackers can successfully deceive systems into verifying against multiple individuals with a single identity. This poses a significant security risk in sensitive domains like border control, where accurate identity verification is crucial. 

In the context of border control, although the European Union has established guidelines on live capture during the enrolment process~\cite{EU-MADRegulations-2019}, many European countries continue to allow passport applicants to submit a previously captured and printed single photograph instead. In case this photograph is a morphed image, it becomes the reference image stored in the passport database. During border control, a live facial capture of the traveller, from a trusted capture device and therefore ensuring to be bona fide, is compared with the reference image in the passport. The morphed reference image can match the two persons who have contributed to the morph, allowing the two individuals to cross borders undetected.

The detection and mitigation of morphing attacks are addressed through two main tasks: single (S-MAD) and differential morphing attack detection (D-MAD). While S-MAD ensures that compromised biometric templates created from a single morphed photograph (e.g. those submitted for official documents such as passports) do not enter the system, thus safeguarding the integrity of the enrolment process, D-MAD looks for identification discrepancies indicative of morphing attacks at the time of identity verification (e.g. at border control) by comparing the live capture with the reference image stored in the passport (system). Algorithms for S-MAD focused mainly on the analysis of PAD-like textural inconsistencies through handcrafted approaches~\cite{Scherhag-MADMultiAlg-ICBEA-2018,Raja-MADBenchmark-TIFS-2020,Dargaud-PCASMAD-WCACV-2023}, image quality degradation~\cite{Debiasi-PRNUVarianceMAD-BTAS-2018,Scherhag-PRNU-TBIOM-2019,Raja-MADBenchmark-TIFS-2020}, pixel discontinuities through noise pattern analysis~\cite{Venkatesh-MADDeepColorResidualNoise-IPTA-2019,Venkatesh-MADResidualNoise-WCACV-2020}, deep features learned by deep neural networks (DNN)~\cite{Zhang-GenSMADDeepRep-CVPR-2024,Caldeira-MADtion-WACV-2025,Tapia-SMADFewShot-Neurocomputing-2025} and hybrid approaches combining multiple feature extractors and classifiers~\cite{Scherhag-MADMultiAlg-ICBEA-2018,Seibold-DeepSMAD-2017}.    

D-MAD algorithms can be broadly classified into two main groups according to~\cite{Venkatesh-MADSurvey-TTS-2021}. The former category includes feature-difference-based approaches, which compare feature vectors of the suspected morphing image and a bona fide image captured in a trusted environment to spot morphing attacks. Numerous approaches have been proposed, such as texture analysis~\cite{Ramachandra-TexturalMAD-ISBA-2019}, 3D gradient~\cite{Mohan-DMAD3DShape-SITIS-2019}, landmark points~\cite{Scherhag-LandmarkMAD-ICISP-2018,Damer-DMADFacialLandmark-PR-2019}, multispectral~\cite{Ramachandra-MultispectralDMAD-WCACV-2024} and deep features~\cite{Damer-DMADMultiDetector-FUSION-2019,Scherhag-FaceMorphingAttacks-TIFS-2020,Ibsen-DifferentialAnomalyDetectionForFacialImages-WIFS-2021}, the latter being the best performing. Most studies focus on digital images, though recent work has improved results using a print and scan dataset~\cite{Scherhag-FaceMorphingAttacks-TIFS-2020,Ngan-MADFRVT-NIST-2020,Tapia-DMADPrintScan-IEEEAccess-2025}. The second group of D-MAD methods are the so-called demorphing techniques, whose aim is to reverse the morphing process in order to discover the original images~\cite{Ferrara-FaceDemorphing-TIFS-2018}. Initially designed for landmark-based morph generation~\cite{Ferrara-FaceDemorphing-TIFS-2018}, recent advancements have primarily utilized DNNs~\cite{Peng-FDGAN-IEEEAccess-2019,Ortega-BorderDeMAD-IEEEAccess-2020}. 

Despite advances reported in the literature showing improved performance of PAD and MAD approaches in unseen target domains, detection pipelines depend on the availability of labelled data from various sources, which is difficult to satisfy in practice (see database summary in Tab~\ref{tab:DB}).  Due to privacy concerns in biometric data acquisition, PAD and MAD algorithms are trained on small databases containing a limited number of domains, resulting in a lack of generalisability~\cite{Ozgur-FoundPAD-ArXiv-2025}. Note that Tab~\ref{tab:DB} databases are constrained in terms of PAI species/morphing tools and number of samples, which limits the generalisability of detection schemes.

\begin{figure*}[!t]
\centering
\includegraphics[width=0.9\linewidth]{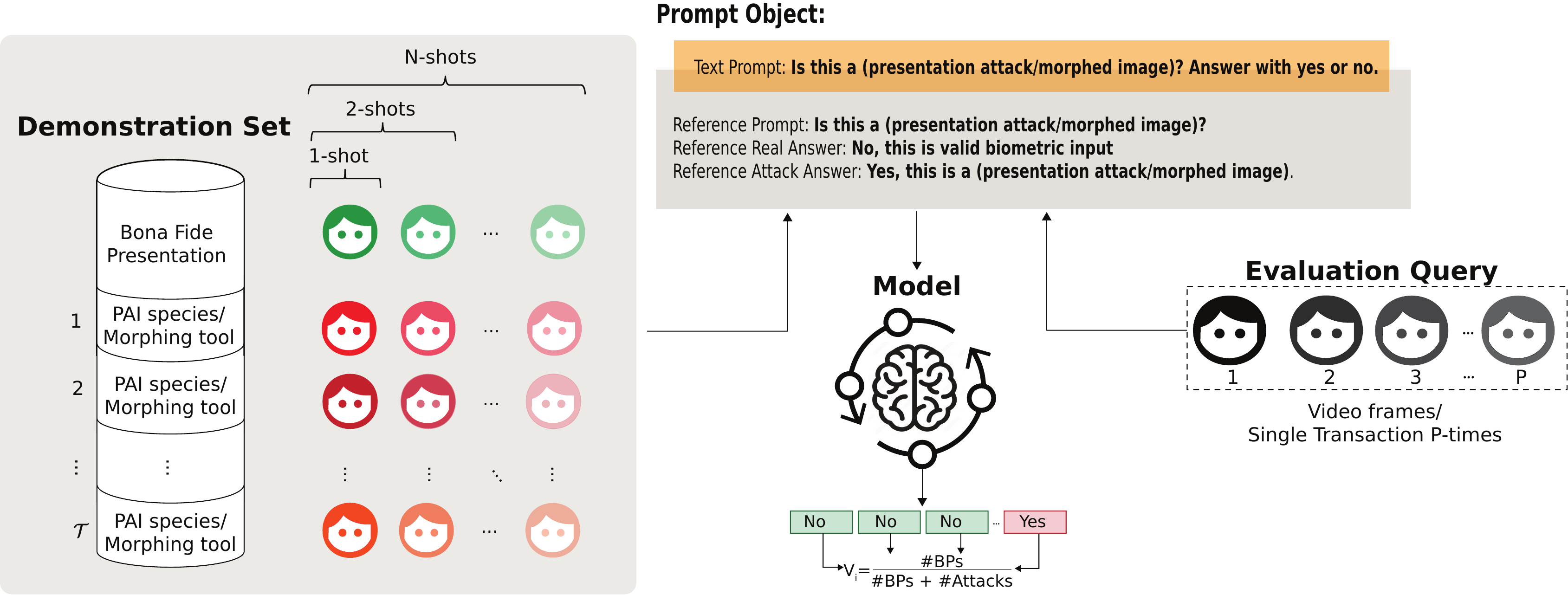}
\caption{Conceptual in-context learning framework for physical (i.e., PAD) and digital (i.e., S-MAD) attack detection.}
\label{fig:overview}
\end{figure*}

\section{Proposed Conceptual In-context Learning Framework}
\label{sec:framework}

Focusing on generalisability, several studies have explored the use of VLMs in biometric and security applications~\cite{Shahreza-FMinBiometrics-ArXiv-2025}. Most of these efforts focus on evaluating the performance of VLMs in recognition tasks, including face recognition~\cite{Hassanpour-ChatGPTFR-ICIP-2024,Deandres-ChatGPTSoftBio-IEEEAccess-2024}, soft biometric estimation~\cite{Hassanpour-ChatGPTFR-ICIP-2024,Deandres-ChatGPTSoftBio-IEEEAccess-2024}, iris recognition~\cite{Farmanifard-ChatGPTIris-IJCB-2024} and gait recognition~\cite{Chivereanu-LLMTextualDesc-FG-2024}. In general, VLMs have demonstrated considerable performance and high generalisation ability in these tasks, which are mainly based on visual appearances. A limited number of approaches have studied the performance of VLMs for PAD~\cite{Shi-SHIELD-ArXiv-2023,Komaty-ChatGPTPAD-ArXiv-2025} and S-MAD~\cite{Zhang-ChatGPTSMAD-2025}.
These analyses are mostly centred on well-known huge VLMs such ChatGPT~\cite{Komaty-ChatGPTPAD-ArXiv-2025,Zhang-ChatGPTSMAD-2025} (GPT-3 has 175 billion parameters~\cite{Mann-LLMsFewLearning-ArXiv-2020}) or Gemini~\cite{Narayan-FaceXBench-ArXiv-2025} (Gemini Pro has over 500 billion parameters\footnote{\url{https://rb.gy/hiarh5}}) and also lack rigorous quantitative evaluation using standardised metrics and were limited to specific attack scenarios. In contrast to previous studies, our work focuses on small, open-source vision-language models (with a maximum of 8 billion parameters), which are lightweight enough for local deployment. This enables privacy-preserving applications by avoiding reliance on server-based processing, a critical consideration in biometric systems.


\subsection{In-context Learning}

While finetuning updates the parameters of the model to adapt it to the task, an alternative approach known as in-context learning allows models to generalise to new tasks without the need to update the parameters. Instead, these models leverage contextual examples within their input to infer task-specific patterns dynamically \cite{Touvron-Llama-ArXiv-2023,Chowdhery-Palm-JMLR-2023}. Intuitively, learning in context lies in learning by analogy. A typical in-context learning pipeline consists of presenting a model with a few demonstration examples formatted as natural language templates, followed by a query~\cite{Wei-ChainPromptLLM-NeurIPS-2022}. By analysing the contextual examples, the model identifies implicit patterns and applies them to generate predictions for the query. This approach eliminates the need for costly retraining, providing a flexible and efficient mechanism for adapting pre-trained models to new targets. In-context learning can be defined as follows~\cite{Mann-LLMsFewLearning-ArXiv-2020}:

Consider a query input \(X\) and a set of candidate answers \(Y = \{y_1, y_2, \dots, y_m\}\). A pretrained language model \(M\) predicts the most likely answer \(\hat{y} \in Y\) by selecting the candidate with the highest conditional probability, given a demonstration set \(C\). The demonstration set \(C\) consists of an optional task instruction \(I\) and \(k\) examples \(\{(x_1, y_1), \dots, (x_k, y_k)\}\), which are formatted to showcase the task. Formally, the likelihood of a candidate answer \(y_i\) is determined by a scoring function \(f_M\):

\[
P(y_i \mid x) = f_M(y_i, C, x).
\]

The predicted answer \(\hat{y}\) is then computed as:

\[
\hat{y} = \operatorname{argmax}_{y_i \in Y} P(y_i \mid x).
\]

Demonstration examples within \(C\) can either follow a task-specific format where all examples belong to the same task or a cross-task format where examples include their respective instructions. The latter enables the model to generalise across diverse tasks~\cite{Dong-InContextLearningSurvey-ArXiv-2022}.

\subsection{Conceptual Framework}

Both PAD and S-MAD are tasks in which the algorithms receive a single image as input and return a confidence score representing the reliance that the input image is a BP~\cite{ISO-IEC-30107-3-PAD-metrics-2023}. Fig.~\ref{fig:overview} shows the in-context learning conceptual framework for PAD and S-MAD. The framework consists of a demonstration set $C$ containing samples of the different classes. Depending on the degree of granularity, the classes can be defined as the different PAI species or morphing tools used in the fabrication of the attacks. In our work, $N$-shots is defined as the number of reference images $N$ selected per category to build the reference prompt for network inference. The reference prompt has the reference real answer for the BP images, while the reference attack answer for all samples is derived from the set of PAI species/morphing tools. The prompt object is then composed of the reference prompt and the text prompt, the latter being used for classification of a given unknown input image (``Evaluation Query'' in Fig.~\ref{fig:overview}). Finally, VLM learns from the reference prompt to answer what it was asked in the text prompt. As the text relies directly on the use of a binary response scheme, we compute the final score for a $P-$frame video as follows: 

\begin{equation}
    V_i = \frac{\#BPs}{\#BPs + \#Attacks},
\end{equation}

\noindent where $\#BPs$ and $\#Attacks$ represent the number of video frames which were classified by the model as bona fide and attack samples, respectively. In this way, we can avoid hallucinations in the score estimation. In case $P = 1$ (i.e., single image classification), the model is asked $K$ times about the classification of the single input image. In our work, we use $K = 5$ when $P = 1$.    







\begin{figure*}[!t]
    \centering
    \subfloat[OULU-NPU]{\includegraphics[width=0.40\linewidth]{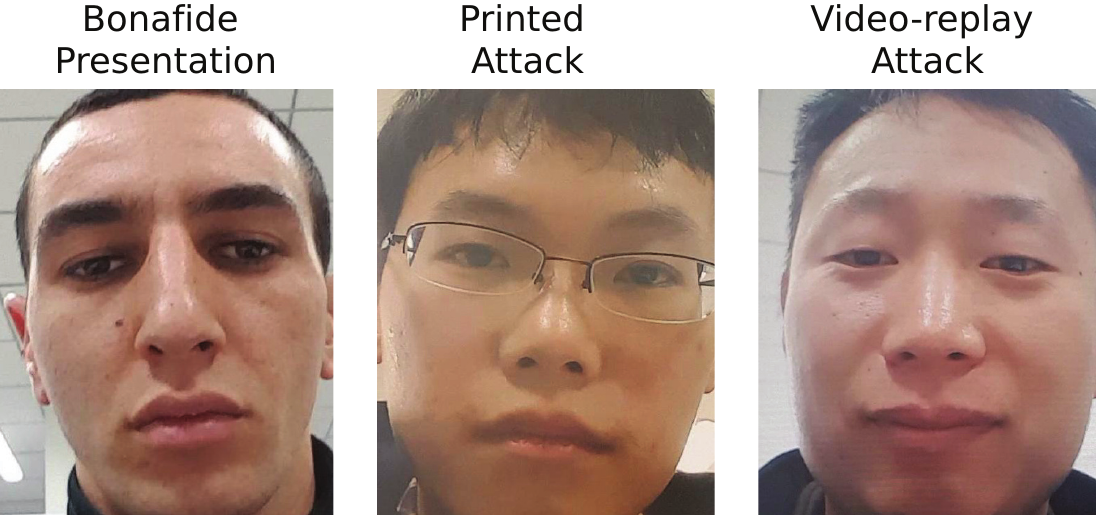}}
    \hfill
    \subfloat[MSU-FASD]{\includegraphics[width=0.40\linewidth]{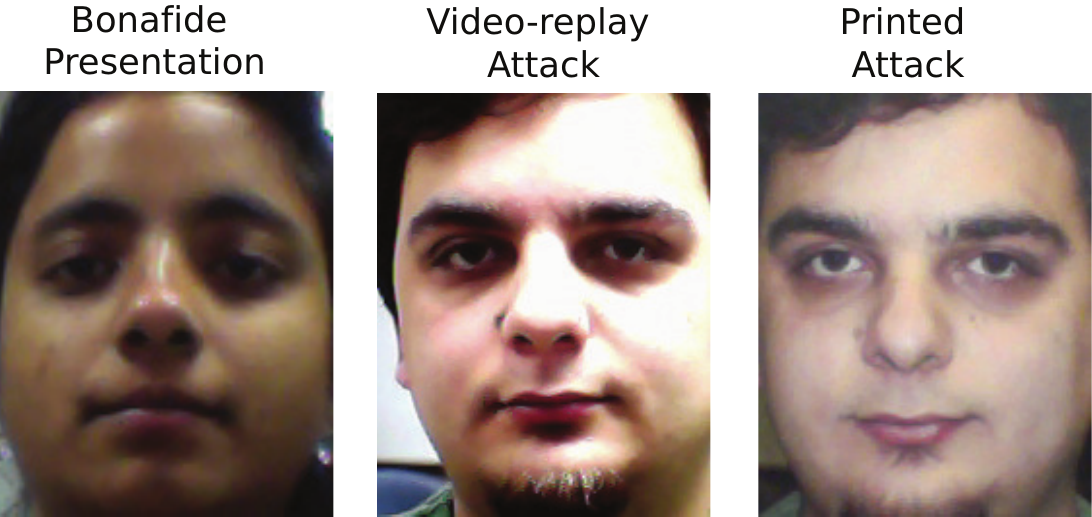}}
    
    \subfloat[CASIA-FASD]{\includegraphics[width=0.45\linewidth]{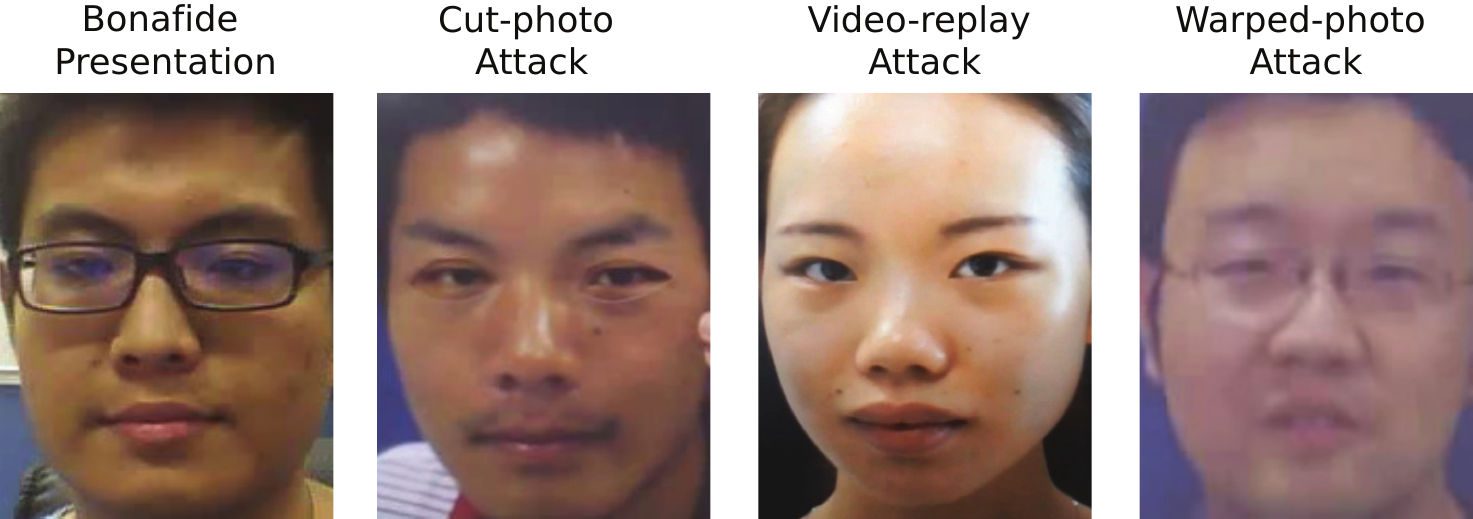}} 
    \hfill
    \subfloat[REPLAY-ATTACK]{\includegraphics[width=0.45\linewidth]{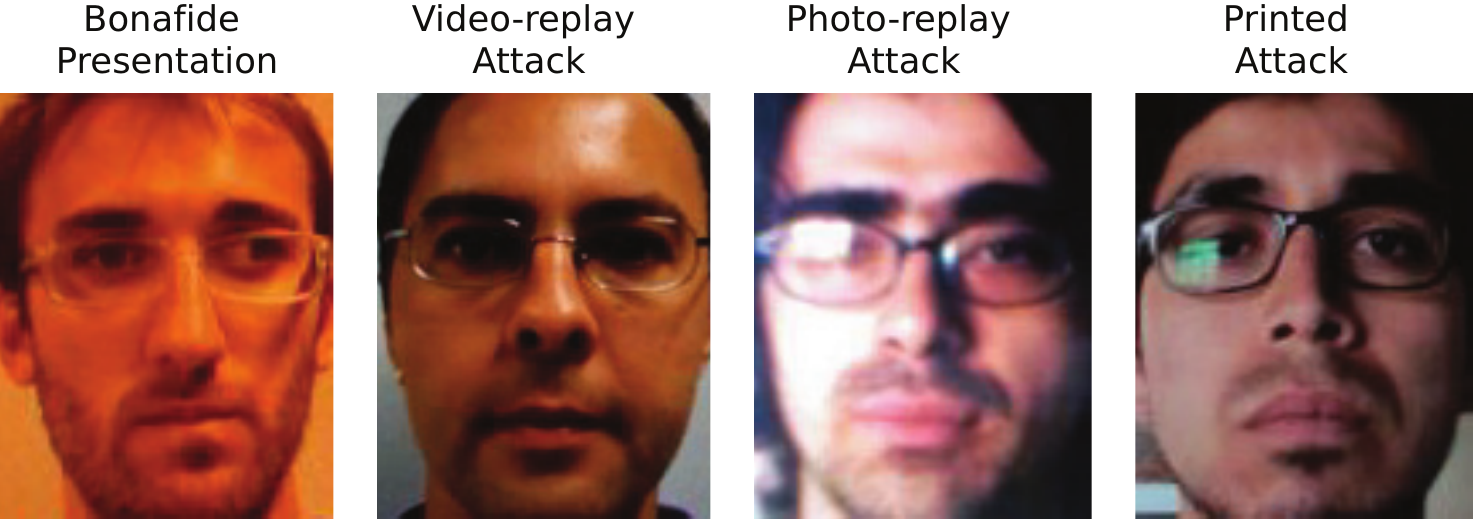}}

    \subfloat[FRGCv2]{\includegraphics[width=0.45\linewidth]{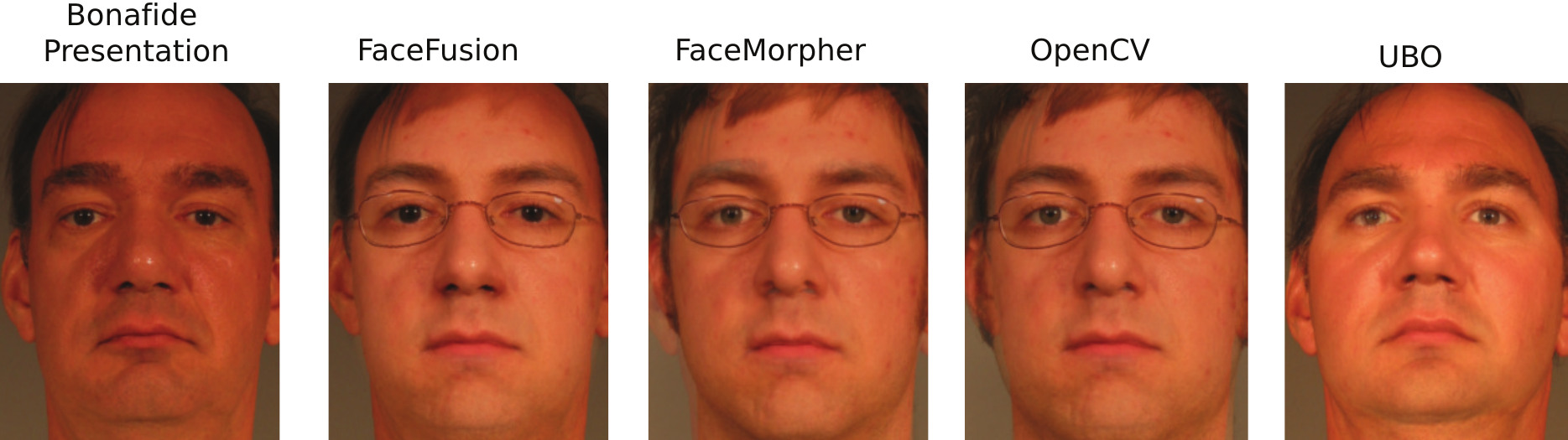}}%
    \hfill
    \subfloat[FERET]{\includegraphics[width=0.45\linewidth]{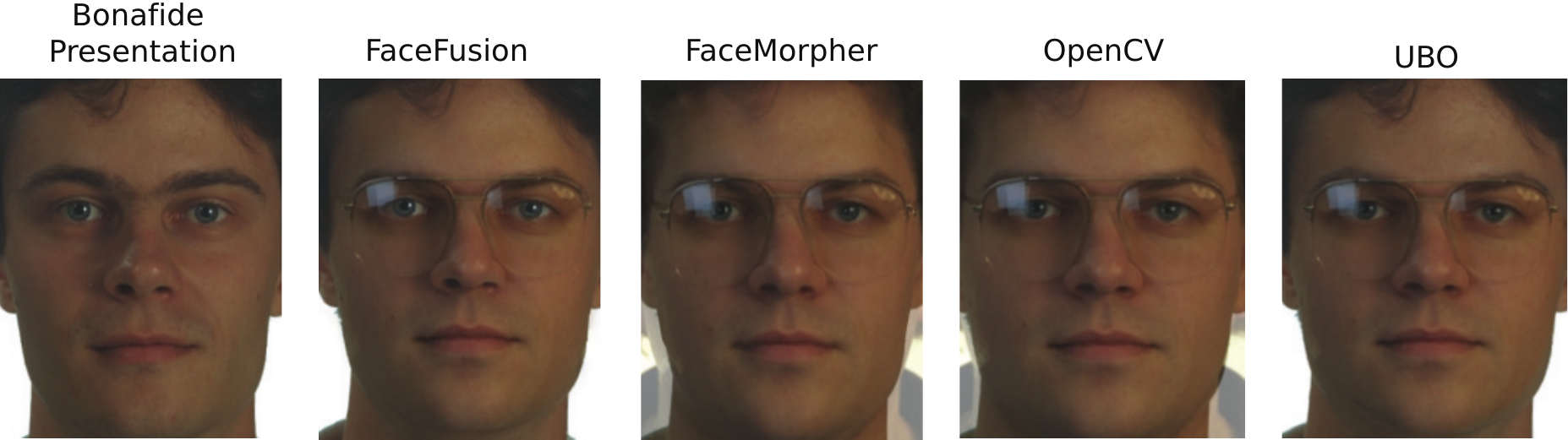}}%
    \caption{Example of BP and PAIs in each database used in the experimental evaluation.}
    \label{fig:databases}
\end{figure*}

\section{Experimental Setup}
\label{sec:exp_setup}

The main goal of the experimental evaluation is the detection performance assessment of the proposed in-context learning framework for zero- and few-shot PAD and S-MAD. To reach our goals, three scenarios are defined:

\begin{itemize}
    \item \textbf{Known‐attacks} scenario reports an analysis of all PAI species. In this scenario, both testing samples and images in the demonstration set were fabricated using the same PAI species.

    \item \textbf{Unknown PAI species} scenario where the PAI species used for testing are different from the PAI species used for the production of the samples in the demonstration set.

    \item \textbf{Cross‐database} is considered the most challenging and realistic, as the datasets used for testing are different (e.g., in terms of subjects, camera, environment conditions, and PAI species) from those used as references in the demonstration set. 
\end{itemize}

\subsection{Databases}

To reach the above goal, the experimental evaluation is carried out on four publicly available databases for PAD: CASIA-FASD~\cite{Zhang-CASIAFASD-ICB-2012}, REPLAY-ATTACK (RA)~\cite{Chingovska-REPLAYATTACK-BIOSIG-2012}, OULU-NPU~\cite{Boulkenafet-OULUNPU-FG-2017} and MSU-FASD~\cite{Wen-MSUFASD-TIFS-2015}. CASIA-FASD~\cite{Zhang-CASIAFASD-ICB-2012} database consists of 600 videos from 50 subjects, including warped-photo, cut-photo and video-replay attacks. REPLAY-ATTACK~\cite{Chingovska-REPLAYATTACK-BIOSIG-2012} contains 1,200 videos from 50 subjects and printed and replay attacks. OULU-NPU~\cite{Boulkenafet-OULUNPU-FG-2017} is a mobile facial PAD dataset, acquired with six different mobile phones and consisting of 4,950 videos from 55 subjects. MSU-FASD~\cite{Wen-MSUFASD-TIFS-2015} dataset includes printed photos and replay attacks, with a total of 440 videos from 35 subjects.

For MAD experiments, FERET~\cite{Phillips-FERET-1998} and FRGCv2~\cite{Phillips-FRGC-2005} databases are considered. The FERET~\cite{Phillips-FERET-1998} database consists of 1,321 BPs and 2,116 MAs, the latter being equally distributed over four morphing tools (i.e., FaceFusion, UBO, FaceMorpher, and OpenCV) that were used for its fabrication~\cite{Scherhag-PRNU-TBIOM-2019}. FRGCv2~\cite{Phillips-FRGC-2005} has 2,710 BP images and 3,856 MA samples, and, like FERET, the MAs were created using the above four morphing tools. Tab.~\ref{tab:DB} summarises the main characteristics of databases and Fig.~\ref{fig:databases} shows examples of BPs and PAIs/MAs for each dataset. 

\subsection{Implementation Details}

As the above PAD databases contain videos, we sampled evenly 5 frames per video across the duration of each video. Subsequently, MTCNN~\cite{Zhang-MTCNN-2016} detects the face per frame, and the resulting image is resized to 224~$\times$~224 pixels. For MAD, face images are cropped as in~\cite{Damer-PrivacySyntheticMAD-CVPR-2022} and, like in PAD, they are resized to 224~$\times$~224 pixels. In the case of MAD, images in the demonstration set are picked from the database that is not being evaluated, e.g. FERET if FRGCv2 is being evaluated. The framework's implementation is based on the Hugging Face~\cite{Jain2-HuggingFace-2022} and PyTorch~\cite{Paszke-PyTorchAnImperative-2019} platforms, which facilitate model choice and setup. Up to 9 shots are evaluated in most experiments. A high number of shots higher than 9 makes computing the inference of current VLMs unfeasible using an 80 GB-DRAM Nvidia A100 GPU.  

\subsection{Model Selection}

\begin{figure*}[!t]
\centering
\includegraphics[width=\linewidth]{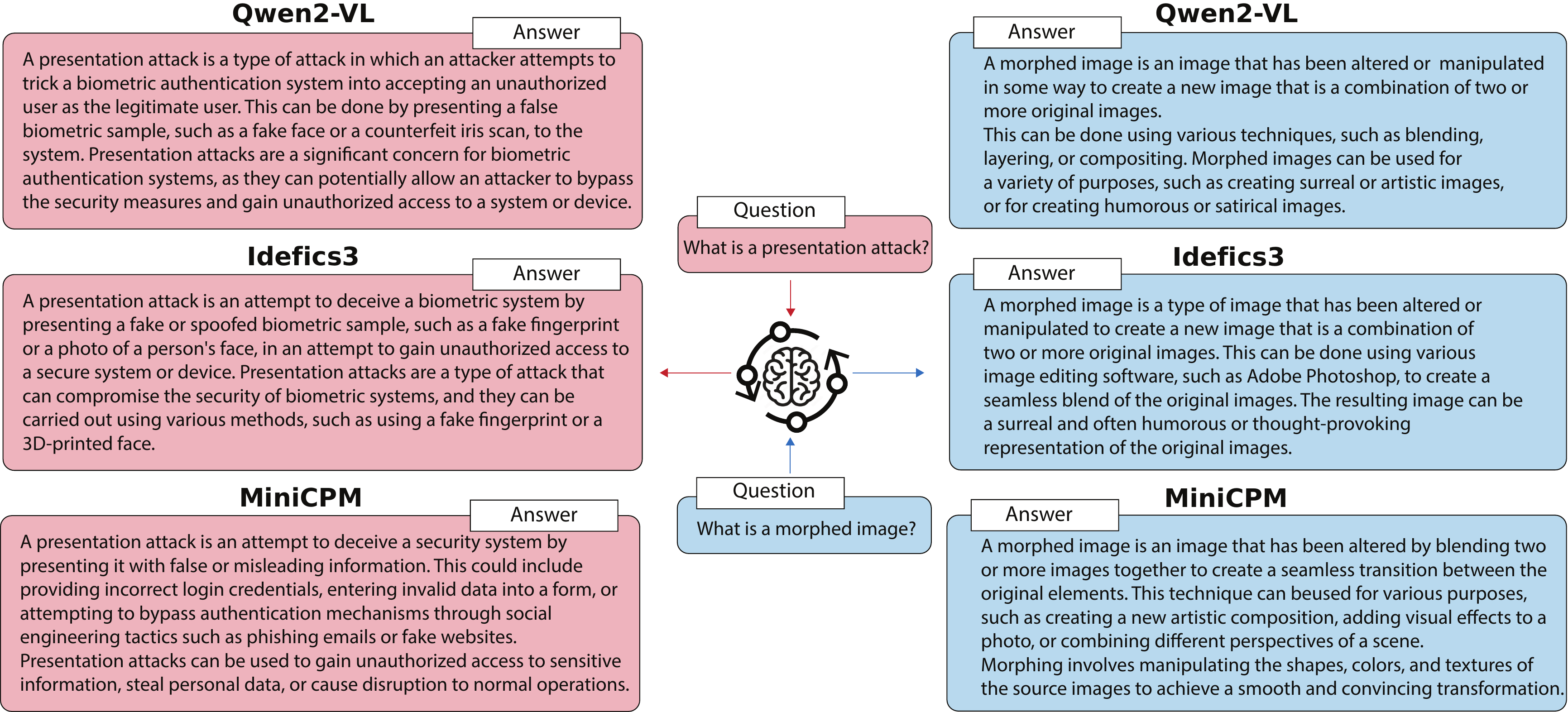}
\caption{Questions on the concepts of presentation (in red) and morphing (in blue) attacks, together with the respective answer provided by the models.}
\label{fig:model_answer}
\end{figure*}

The model selection criteria give priority to efficient open-source models. In comparison to the state-of-the-art (SOTA) models like GPT-4o and Gemini using over 200 billion parameters, this work focuses on practical efficiency, testing models within 8 billion parameters. Due to the compact, 20 times smaller size, those models might be suitable for real-world applications. Based on the Hugging Face Open Leaderboard~\cite{Duan-Vlmevalkit-ACM-2024}, three model families have been selected: Qwen2-VL, Idefics3 and MiniCPM. All models are able to understand the concepts of presentation and morphing attacks, as shown in Fig.~\ref{fig:model_answer}. Our work has also evaluated other families of models (e.g., Ovis2 \cite{ex-s-ovis}, and InternVL \cite{ex-s-intern}), all of which exceeded the capabilities of the available hardware.  

Qwen2-VL\footnote{\url{https://huggingface.co/Qwen/Qwen2-VL-7B-Instruct}} is a SOTA multimodal language model developed by Alibaba Cloud's Qwen team~\cite{ex-s-qwen-1, ex-s-qwen-2}, part of the Qwen2 series. The model's architecture is built on a dense transformer with 7 billion parameters. According to multiple visual-task specific benchmarks \cite{ex-s-mme, ex-s-ocrbench, ex-s-mmstar}, the model excels in understanding visual elements at a comparable level to GPT-4o-mini~\cite{ex-s-qwen}. Therefore, the model is expected to recognise not only faces, but also to interpret associated information, such as presentation attack instruments or contextual clues.


Idefics3\footnote{\url{https://huggingface.co/HuggingFaceM4/Idefics3-8B-Llama3}} was selected for comprehensive evaluation due to its strong performance on key vision-language benchmarks, most notably a +13.7 point leap on DocVQA—underscoring its enhanced OCR, document comprehension, and reasoning capabilities. Similar to MiniCPM, Idefics3 comprises 8 billion parameters and achieves notably improved visual reasoning and document understanding compared to its predecessor, Idefics2, as it integrates an advanced Vision-Language Architecture. Idefics3 combines a SigLIP‑SO400M image encoder with the Llama 3 language model, replacing Idefics2’s perceiver and introducing an updated image-processing logic—including a pixel-shuffle strategy that compresses visual input into 169 tokens via a 364~$\times$~364 patch grid with positional cues—thereby boosting efficiency without sacrificing structure.    

MiniCPM-V 2.6\footnote{\url{https://huggingface.co/openbmb/MiniCPM-V-2_6}} represents a SOTA omnimodal architecture featuring 8 billion parameters with integrated multimodal encoders and decoders trained through end-to-end optimisation \cite{ex-s-mini}. This model demonstrates exceptional processing efficiency through its token compression mechanism - analysis of 1.8 megapixel images requires only 640 tokens, representing a 75\% reduction compared to GPT-4o's tokenisation approach \cite{ex-s-mini-v}. 


\subsection{Evaluation Metrics}

The experimental results are analysed and reported in compliance with the metrics defined in the International Standards ISO/IEC 30107-3~\cite{ISO-IEC-30107-3-PAD-metrics-2023} for biometric PAD and ISO/IEC 20059~\cite{ISO-IEC-20059} for MAD:

\begin{itemize}
    \item Attack Presentation/Morphing Attack Classification Error Rate (APCER/MACER), which computes the proportion of attack presentations/morphing attacks wrongly classified as bona fide presentations.

    \item Bona Fide Presentation/Sample Classification Error Rate (BPCER/BSCER), which is defined as the proportion of bona fide presentations misclassified as attack presentations (morphing samples).
\end{itemize}

Based on these metrics, we report $i)$ the BPCERs/BSCERs observed at APCER/MACER values or security thresholds of 1\% (BPCER/BSCER100), 5\% (BPCER/BSCER20), and 10\% (BPCER/BSCER10); and $ii)$ the Detection Equal Error Rate (D-EER), which is defined as the error rate value at the operating point where APCER=BPCER / MACER=BSCER. To benchmark against the state of the art, non-ISO compliant metrics are also presented, i.e., Half-Total Error Rate (HTER) and Area Under the Receiver Operating Characteristic (ROC) Curve (AUC).

\section{Results and Discussion}
\label{sec:results}

The experimental results are presented taking into account the scenarios defined in Sect.~\ref{sec:exp_setup}. While known-attack scenarios are evaluated in Sect.~\ref{sec:known_Attacks}, Sect.~\ref{sec:unknown_Attacks} and Sect.~\ref{sec:cross_db} report an in-depth performance analysis for unknown PAI species and cross-database scenarios, respectively. Sect.~\ref{sec:benchmarkSOTA} provides a benchmark of the proposed framework against the state-of-the-art for zero-shot PAD and S-MAD.  

\subsection{Known-attacks}
\label{sec:known_Attacks}

\begin{table*}[!t]
    \centering
    \caption{Detection performance (in \%) of VLMs for known-attacks scenarios on CASIA-FASD.}
    \label{tab:KA-CASIA}
    \begin{tabular}{c|r|r|c|c c c c}
    \toprule \toprule
        \textbf{Model} & \textbf{References} & \textbf{Testing} & \textbf{Shots} & \textbf{D-EER} & \textbf{BPCER10} & \textbf{BPCER20} & \textbf{BPCER100} \\ 
    \midrule \midrule
\multirow{3}{*}{Idefics3} & cut\_attack & cut\_attack & 9 & 33.89 & 57.78 & 76.67 & 88.89 \\ 
                           & cut\_attack, warped\_attack & warped\_attack & 1 & 32.22 & 75.56 & 75.56 & 75.56 \\ 
                           & warped\_attack, video\_attack & video\_attack & 5 & 44.44 & 81.11 & 87.78 & 88.89 \\ 
\midrule
\multirow{3}{*}{MiniCPM}  & cut\_attack & cut\_attack & 7 & 16.11 & 36.67 & 54.44 & 54.44 \\ 
                          & cut\_attack, warped\_attack, video\_attack & warped\_attack & 5 & 25.00 & 56.67 & 80.00 & 93.33 \\ 
                          & video\_attack & video\_attack & 7 & 30.00 & 65.56 & 81.11 & 93.33 \\ 
                           
\midrule
  \multirow{3}{*}{Qwen2} & cut\_attack, warped\_attack & cut\_attack & 3 & \textbf{11.67} & \textbf{12.22} & \textbf{18.89} & \textbf{41.11} \\ 
                         & warped\_attack & warped\_attack & 3 & \textbf{22.78} & \textbf{58.89} & \textbf{77.78} & \textbf{77.78} \\ 
                         & video\_attack & video\_attack & 1 & \textbf{26.11} & \textbf{44.44} & \textbf{44.44} & \textbf{72.22} \\ 
\bottomrule \bottomrule
    \end{tabular}
\end{table*}

\begin{figure}[!t]
\centering
\includegraphics[width=0.9\linewidth]{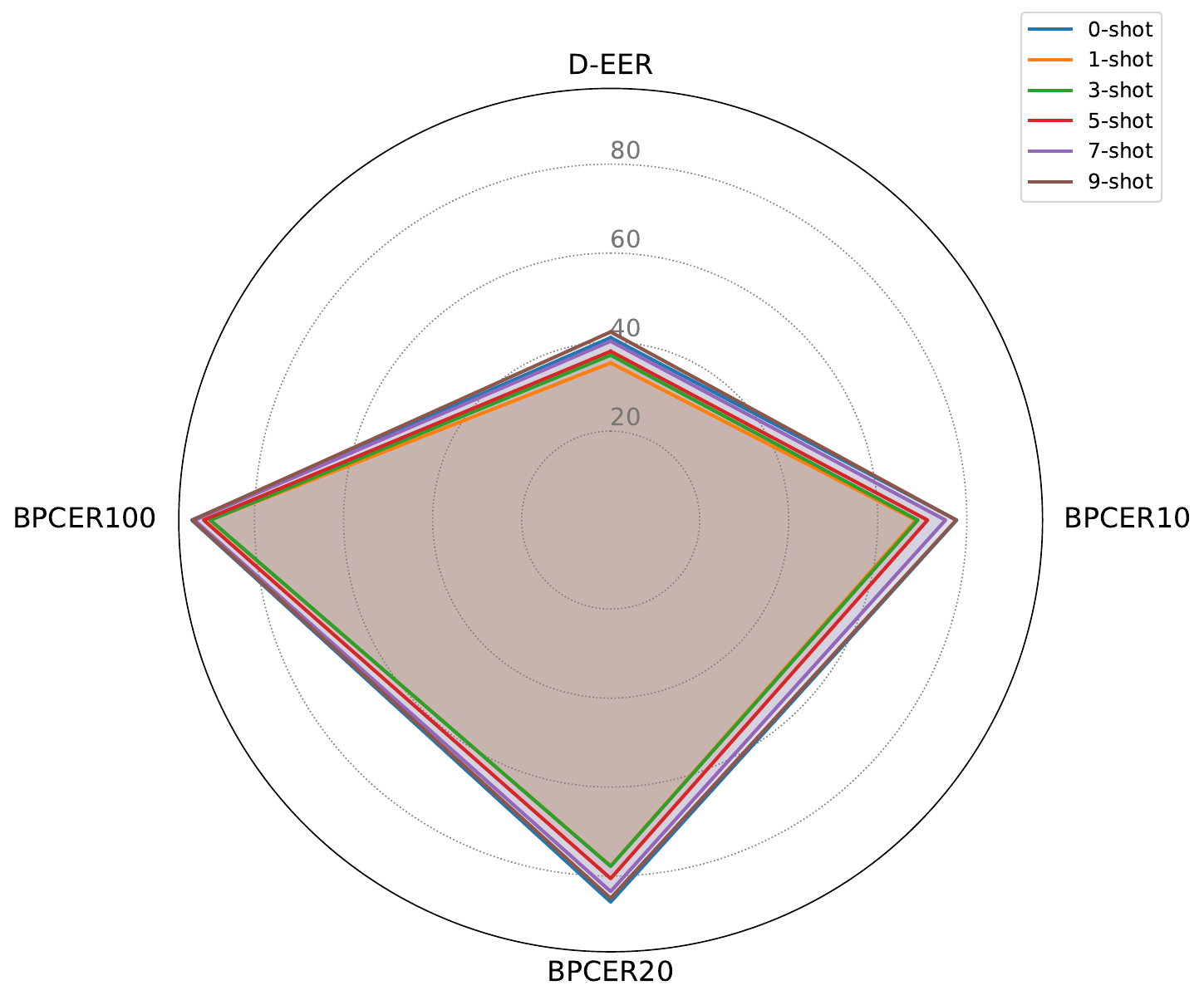}
\caption{Performance trends worsen with the number of shots. }
\label{fig:perf_trends_avg.}
\end{figure}

\begin{table*}[!t]
    \centering
    \caption{Detection performance (in \%) of VLMs for unknown-PAI scenarios on CASIA-FASD.}
    \label{tab:UA-CASIA}
    \begin{tabular}{c|r|r|c|c c c c}
    \toprule \toprule
        \textbf{Model} & \textbf{References} & \textbf{Testing} & \textbf{Shots} & \textbf{D-EER} & \textbf{BPCER10} & \textbf{BPCER20} & \textbf{BPCER100} \\ 
    \midrule \midrule
\multirow{3}{*}{Idefics3} & warped\_attack, video\_attack & cut\_attack     & 1 & 37.22	&   70.00  &  70.00	 &  70.00 \\ 
                          & cut\_attack                  & warped\_attack  & 9 & 38.89 &   76.67	&  92.22  &  92.22 \\ 
                          & cut\_attack, warped\_attack  & video\_attack   & 1 & 40.56 &   78.89	&  78.89  &  78.89 \\ 
\midrule
\multirow{3}{*}{MiniCPM}  & warped\_attack & cut\_attack & 7 & 20.56  &  42.22	&  72.22	&  92.22 \\ 
                          & cut\_attack    & warped\_attack & 3 & 25.56 &  34.44	&  34.44  &	 34.44 \\ 
                          & cut\_attack    & video\_attack & 5 & 29.44  & 51.11 & 51.11 & 51.11 \\ 
                           
\midrule
  \multirow{3}{*}{Qwen2} &      -         & cut\_attack     & 0 & \textbf{10.56} & \textbf{12.22} & \textbf{30.00} &   \textbf{43.33} \\
                         & cut\_attack    & warped\_attack  & 9 & \textbf{17.22} & \textbf{24.44} & \textbf{41.11} &   \textbf{76.67} \\ 
                         & warped\_attack & video\_attack   & 7 & \textbf{25.56} & \textbf{47.78} & \textbf{47.78} &   \textbf{78.89} \\ 
                          
\bottomrule \bottomrule
    \end{tabular}
\end{table*}

\begin{figure*}[!t]
\centering
\subfloat[CASIA - MSU]{\includegraphics[width=0.25\linewidth]{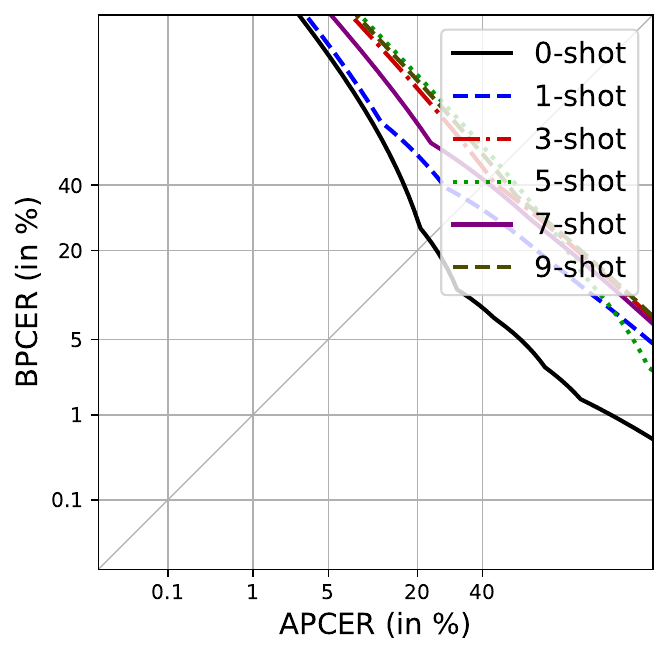}}%
\hfill
\subfloat[OULU - RA]{\includegraphics[width=0.25\linewidth]{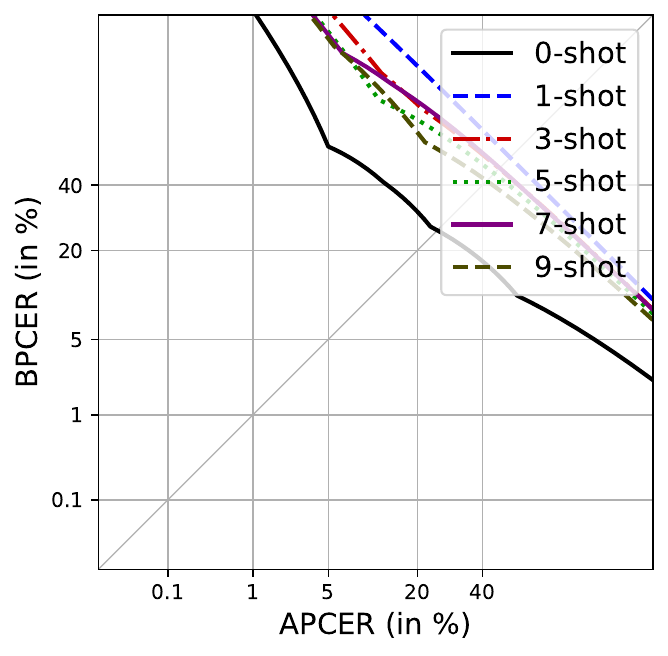}}%
\hfil
\subfloat[MSU - CASIA]{\includegraphics[width=0.25\linewidth]{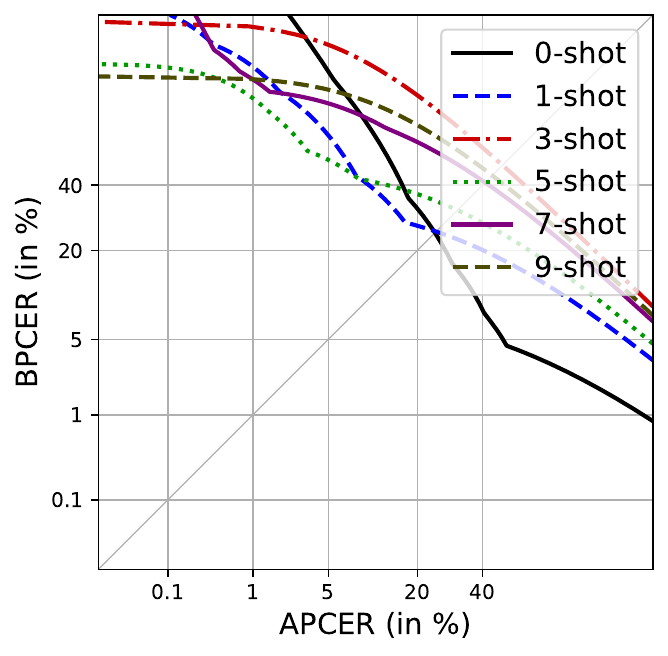}}%
\hfill
\subfloat[RA - OULU]{\includegraphics[width=0.25\linewidth]{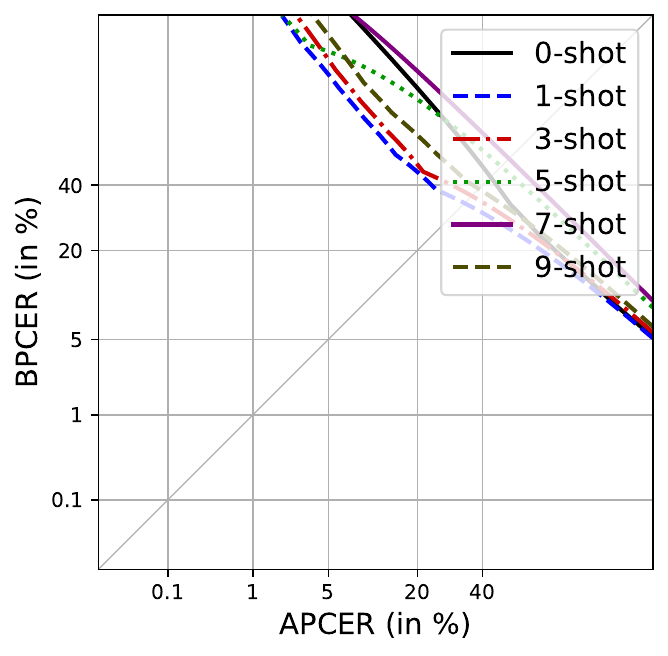}}%
\caption{Cross-database performance of Qwen2 plotted as DET curves for different shots.}
\label{fig:CD}
\end{figure*}

The framework's performance using different models for known-attacks scenarios is computed on CASIA-FASD and reported in Tab.~\ref{tab:KA-CASIA}. To do so, we evaluate all PAI species combinations in the demonstration set selected from the CASIA-FASD training set and show the best performers per test PAI species. In line with the known-attack settings, the test PAI species are always in the demonstration set. Note that Qwen2 achieves the best performance using only a few samples in the demonstration set: at most 3 reference samples per category are enough for Qwen2 to yield D-EERs in the range 11\%-26\%. MiniCPM also offers similar detection performance to Qwen2, but considering more samples in the demonstration set, while Idefics3 is the worst performer. For higher security thresholds (i.e., BPCER100), the best performance is dominated by Qwen2, which reports a BPCER100 in the range 41\%-77\%. 


Since the models perform differently depending on the number of shots, we investigated the average trend of the VLMs' performance as the number of shots increases. For this purpose, we average independently each metric (i.e., D-EER, BPCER10, BPCER20, and BPCER100) for all models per shot and plot the results in  Fig.~\ref{fig:perf_trends_avg.}. Note that, except for zero-shot, all operating points (D-EER, BPCER10, BPCER20, and BPCER100) from one-shot onwards get worse with the number of shots, thus confirming the same findings reported by~\cite{Huang-ICLReasoning-ArXiv-2025}: VLMs' performance can plateau or even degrade for a high number of shots due to factors as context window limitations, visual–textual interference, and lack of instruction tuning.

\subsection{Unknown-attacks}
\label{sec:unknown_Attacks}

Tab.~\ref{tab:UA-CASIA} reports the performance of our framework for different VLMs on unknown PAI species scenarios from CASIA-FASD, i.e.,  the test PAI species are unknown in the demonstration set. Note that Qwen2 shows the best generalisation capability, resulting in performance similar to the one in Tab.~\ref{tab:KA-CASIA}. Compared to the known attack scenarios, the D-EER and BPCER values for different operating points in unknown attack scenarios improve, especially for warped\_attacks, resulting in an enhancement in terms of D-EER of almost 6 percentage points (i.e., 17.22\% vs. 22.78\%). Contrary to the known attacks, we observe that the results are mostly achieved on a large number of shots for all models. We can also see that the use of cut\_attack in the demonstration set allows the efficient detection of most of the unknown PAI species. With the exception of video\_attack detection by Qwen2, cut\_attack appears to be the most suitable PAI species for achieving high generalisability in the detection of other PAI species. Unlike traditional supervised learning approaches, which rely on the strict use of labelled data, our in-context learning framework learns patterns by analogy from a few shots of the demonstration set to classify unknown samples, leading to high generalizability. The results indicate that the in-context learning framework does not overfit the demonstration set, unlike traditional supervised learning approaches, which perform better when training and test data are produced using the same set of PAI species.

As Qwen2 reports the best performance for known and unknown attack scenarios, it is selected and evaluated alone for the rest of the protocols.   

\subsection{Cross-database}
\label{sec:cross_db}

\begin{table*}[!t]
    \centering
    \caption{Detection performance (in \%) of Qwen2 for cross-database scenarios on FERET and FRGC. The best results per training-testing database combination and testing morphing tool are highlighted in bold.}
    \label{tab:CD-FERET-FRGC}
    \begin{adjustbox}{width=\linewidth}
    \begin{tabular}{c|r|r|c c c c| c c c c}
    \toprule \toprule
    \multirow{2}{*}{\textbf{Databases}} & \multirow{2}{*}{\textbf{References}} & \multirow{2}{*}{\textbf{Testing}} & \multicolumn{4}{c|}{\textbf{Cropped Faces}} & \multicolumn{4}{c}{\textbf{Uncropped Faces}}  \\ 
                             &  &  & \textbf{Shots} & \textbf{D-EER} & \textbf{BSCER10} & \textbf{BSCER20} & \textbf{Shots} & \textbf{D-EER} & \textbf{BSCER10} & \textbf{BSCER20} \\ 
    \midrule \midrule
\multirow{16}{*}{FERET-FRGC} & \multirow{4}{*}{morphs\_facefusion}  & morphs\_facefusion    & 1 & 48.58 &  89.61    &  94.81     &  5  & \textbf{33.06} &  \textbf{79.89}    &   \textbf{89.95} \\ 
                            &                                       & morphs\_facemorpher   & 5 & 39.77 &  85.01    &  92.51  	 &  1  & 12.51 &  53.86    & 76.93 \\  
                            &                                       & morphs\_opencv        & 1 & 41.82 &  87.76    &  93.88  	 &  3 & 21.58 &  73.04    &  86.52   \\ 
                            &                                       & morphs\_ubo           & 1 & 46.62 &  83.80    &  91.90  	 &  5 & 45.54 &  84.45    &  92.23   \\ 
                            \cmidrule{2-11}
                            & \multirow{4}{*}{morphs\_facemorpher}  & morphs\_facefusion    & 3 & 42.18 &  84.47    &  92.24  	 & 1 & 41.55 &  87.71    &  93.86   \\ 
                            &                                       & morphs\_facemorpher   & 3 & 37.73 &  82.99    &  91.49  	 & 3 & 13.61 &  59.49    &  79.75   \\  
                            &                                       & morphs\_opencv        & 1 & 44.62 &  88.27    &  94.13  	 & 5 & \textbf{15.42} &  \textbf{49.09}    &  \textbf{74.55}   \\ 
                            &                                       & morphs\_ubo           & 3 & 45.75 &  88.46    &  94.23  	 & 7 & 44.78 &  87.44    &  93.72  \\ 
                            \cmidrule{2-11}
                            & \multirow{4}{*}{morphs\_opencv}       & morphs\_facefusion    & 7 & 49.00 &  88.10    &  93.04  	 &  1 & 45.35 &  88.94    &  94.47  \\   
                            &                                       & morphs\_facemorpher   & 1 & 35.39 &  84.72    &  92.36  	 &  5 & 12.46 &  53.81    &  76.90  \\  
                            &                                       & morphs\_opencv        & 1 & 40.92 &  86.98    &  93.49  	 &  3 & 16.56 &  67.05    &  83.53   \\
                            &                                       & morphs\_ubo           & 1 & 45.01 &  88.75    &  94.37  	 &  5 & 44.67 &  86.96    &  93.48  \\
                            \cmidrule{2-11}
                            & \multirow{4}{*}{morphs\_ubo}          & morphs\_facefusion    & 5 & 48.77 &  88.45    &  94.23  	 &  5 & 40.42 &  78.94    &  89.47  	  \\
                            &                                       & morphs\_facemorpher   & 1 & 39.87 &  87.07    &  93.54  	 &  1 & \textbf{12.54} &  \textbf{19.58}    &  \textbf{20.67}  	  \\
                            &                                       & morphs\_opencv        & 1 & 40.40 &  85.04    &  92.52  	 &  3 & 19.15 &  58.86    &  79.43  	 \\ 
                            &                                       & morphs\_ubo           & 5 & \textbf{43.45} &  \textbf{81.25}    &  \textbf{90.62}  	 &  5 & 46.38 &  84.75    &  92.38  \\
\midrule
\multirow{16}{*}{FRGC-FERET}& \multirow{4}{*}{morphs\_facefusion}   & morphs\_facefusion    & 1 & 48.24 &  89.37    &  94.68  	 &  1 & \textbf{39.60} &  \textbf{79.22}    &  \textbf{89.61}  \\
                            &                                       & morphs\_facemorpher   & 5 & 36.90 &  65.39    &  82.70  	 &  0 & \textbf{13.89} &  \textbf{56.49}    &  \textbf{78.25}  	  \\ 
                            &                                       & morphs\_opencv        & 1 & 40.92 &  80.80    &  90.40  	 &  0 & \textbf{9.92}  &  \textbf{31.11}    &  \textbf{65.56}  	  \\
                            &                                       & morphs\_ubo           & 1 & 48.69 &  89.73    &  94.86  	 &  1 & 44.90 &  87.97    &  93.98  	  \\
                            \cmidrule{2-11}
                            & \multirow{4}{*}{morphs\_facemorpher}  & morphs\_facefusion    & 7 & 48.42 &  86.85    &  93.35  	 &  1 & 45.75 &  87.98    &  93.99    \\ 
                            &                                       & morphs\_facemorpher   & 1 & 37.06 &  86.17    &  93.09  	 &  0 & \textbf{13.89} &  \textbf{56.49}    &  \textbf{78.25}  	  \\ 
                            &                                       & morphs\_opencv        & 5 & 39.24 &  80.32    &  90.16  	 &  0 & \textbf{9.92}  &  \textbf{31.11}    &  \textbf{65.56}  	  \\
                            &                                       & morphs\_ubo           & 1 & \textbf{42.15} &  86.98    &  93.49  	 &  1 & 42.91 &  87.49    &  93.75  	  \\
                            \cmidrule{2-11}
                            & \multirow{4}{*}{morphs\_opencv}       & morphs\_facefusion    & 1 & 48.22 &  87.49    &  93.75  	 &  3 & 45.46  &  86.78   & 93.39 \\    
                            &                                       & morphs\_facemorpher   & 1 & 44.97 &  82.43    &  91.22  	 &  0 & \textbf{13.89} &  \textbf{56.49}    &  \textbf{78.25}  	  \\  
                            &                                       & morphs\_opencv        & 5 & 46.14 &  82.32    &  90.91  	 &  0 & \textbf{9.92}  &  \textbf{31.11}    &  \textbf{65.56}  	  \\ 
                            &                                       & morphs\_ubo           & 1 & 45.78 &  87.75    &  93.87  	 &  3 & 43.29 &  86.60    &  93.40 \\  
                            \cmidrule{2-11}
                            & \multirow{4}{*}{morphs\_ubo}          & morphs\_facefusion    & 1 & 47.24 &  88.56    &  94.28  	 &  1 & 43.95 &  84.29    &  92.14 \\  
                            &                                       & morphs\_facemorpher   & 1 & 46.95 &  87.60    &  93.80  	 &  0 & \textbf{13.89} &  \textbf{56.49}    &  \textbf{78.25}  	  \\ 
                            &                                       & morphs\_opencv        & 5 & 48.26 &  86.78    &  91.61  	 &  0 & \textbf{9.92}  &  \textbf{31.11}    &  \textbf{65.56}  	  \\ 
                            &                                       & morphs\_ubo           & 1 & 46.88 &  88.33    &  94.16  	 &  1 & 42.91 &  \textbf{86.03}    &  \textbf{93.02} \\  
\bottomrule \bottomrule
    \end{tabular}
    \end{adjustbox}
\end{table*}

The development of PAD subsystems has quickly evolved over the years, especially with the development of deep neural networks. Contrary to technological progress, the creation of new databases to train and achieve generalisability of such algorithms is slower due to certain privacy issues and is a time-consuming task. In real applications, changes in environmental conditions, unknown PAI species and even subjects cause a shift in the statistical distribution of test images and thus poor PAD performance. The cross-database generalizability of the proposed framework using Qwen2 as a backbone is evaluated. For this purpose, all demonstration-test database combinations (i.e., CASIA-MSU, CASIA-OULU, CASIA-RA, etc.) are evaluated, and the best performing ones are shown in Fig.~\ref{fig:CD} as DET curves for different numbers of shots. 

Note that the framework decreases in performance mainly with the number of shots, with the zero-shot being the best performer in terms of D-EER on average. In particular, the D-EER is around 22\% for all test databases, except for OULU-NPU, which exceeds 40\%. This is partly due to the image quality of the OULU, which is superior to the image quality of the other databases. In terms of operating points (i.e., BPCER@APCER~$\leq$~10\%), we observe that the performance is in the range 26\%~$\leq$~BPCER~$\leq$~61\% at an APCER=10\%, demonstrating the generalisability of the proposed framework to perform in unfamiliar environments (i.e., cross database) without expert knowledge of the task (i.e., zero-shot PAD).  

\begin{figure}[!t]
\centering
\includegraphics[width=\linewidth]{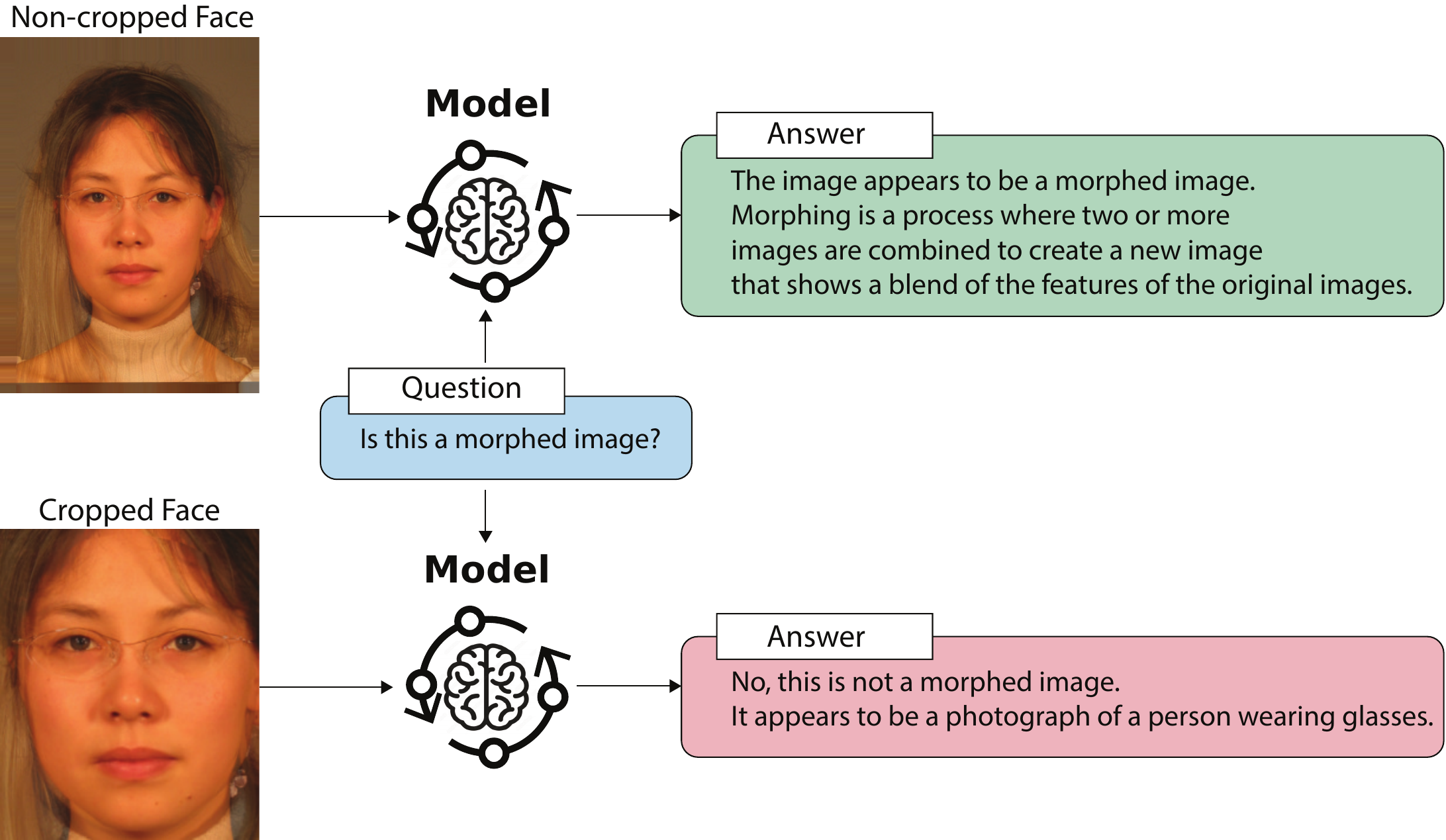}
\caption{Qwen2 response to a morphed image with one uncropped face (top) and one cropped face (bottom).}
\label{fig:SMADbackgroundIssues}
\end{figure}

In addition to the PAD experiments, we investigate the feasibility of the proposed in-context learning framework for S-MAD. To that end, Qwen2 is selected, and a cross-database evaluation is conducted using FERET and FRGC. Tab.~\ref{tab:CD-FERET-FRGC} reports the ISO-compliant metrics summarising the performance of the Qwen2-based framework for different combinations of morphing tools (e.g., morphs\_facefusion from FERET (in the demonstration set) - morphs\_facemorpher from FRGC (in the test set)). We observed during the experiments that the performance of the models for S-MAD decreases significantly due to face clipping. Fig.~\ref{fig:model_answer} shows the Qwen2 response for a morphed FRGC image in which the face was cropped. The network was also asked to give a response for the same face without cropping. Note that the model cannot infer the nature of the image when the face is cropped (answer in the red box). The answer is:

\begin{center}
    ``\textit{No, this is not a morphed image. It appears to be a photograph of a person wearing glasses.}''
\end{center}

However, Qwen2 changed the answer when the full image is provided (answer in the green box):        

\begin{center}
    ``\textit{The image appears to be a morphed image. Morphing is a process where two or more images are combined to create a new image that shows a blend of the features of the original images.}''
\end{center}

Based on the above finding, Tab.~\ref{tab:CD-FERET-FRGC} also reports a comparison of the Qwen2 performance for cropped and uncropped faces. Note that the performance of the frame improves considerably when an uncropped facial image is presented as input. Performance improvements went from 49\% to 9\% in terms of D-EER depending on the morphing tools when using an uncropped face. It is worth noting that morphs\_facemorpher and morphs\_opencv are the easiest attacks to detect, while morphs\_ubo and morphs\_facefusion seem to be the most difficult. While the D-EER values of the first two are between 9\% and 15\%, those of the latter two are above 33\% for uncropped faces. Note the effect of the uncropped faces on the detection of the easiest attacks (i.e. morphs\_facemorpher and morphs\_opencv) in terms of number of shots: the performance (D-EER) of the model went from 40\% to 9\% by decreasing the number of shots from five to zero in many cases. Based on the above results, we strongly believe that the same effect of uncropped faces can be observed in the PAD, which would result in a considerable improvement of the performance presented in the previous sections.        

\subsection{Benchmark with the State-of-the-art}
\label{sec:benchmarkSOTA}

\begin{table*}[!t]
    \centering
    \caption{Benchmark of Qwen2 against the state-of-the-art PAD for zero-shot learning. Performance is reported in percentages, with the best results highlighted in bold.}
    \label{tab:benchmarkSOTA}
    \begin{tabular}{r| c c| c c| c c| c c| c c}
    \toprule \toprule
        \multirow{2}{*}{\textbf{Approaches}} & \multicolumn{2}{c|}{\textbf{MSU}} & \multicolumn{2}{c|}{\textbf{CASIA}}  & \multicolumn{2}{c|}{\textbf{RA}} & \multicolumn{2}{c|}{\textbf{OULU}} & \multicolumn{2}{c}{\textbf{Avg.}} \\ 
                                &   HTER$\downarrow$ & AUC$\uparrow$ &  HTER$\downarrow$ & AUC$\uparrow$ & HTER$\downarrow$ & AUC$\uparrow$  &  HTER$\downarrow$ & AUC$\uparrow$ &  HTER$\downarrow$ & AUC$\uparrow$ \\
    \midrule \midrule
FoundPAD[TI] (Vit-B)~\cite{Ozgur-FoundPAD-ArXiv-2025} & 55.71   &   41.22  &  50.67  &  49.53  & 50.50 &  50.74 &  52.05 & 47.87 &  52.23 & 47.34  \\
FoundPAD[TI] (Vit-L)~\cite{Ozgur-FoundPAD-ArXiv-2025} & 41.19   &   62.96  &  43.44  &  56.56  & 46.50 &  54.49 &  44.76 & \textbf{59.44} &  43.97 & 58.36 \\
\midrule
ours (Qwen2)                                          & \textbf{23.40}   &   \textbf{82.37}  &  \textbf{24.07}  &  \textbf{82.09}  & \textbf{24.99} &  \textbf{79.16} & \textbf{41.67}  & 59.16 &  \textbf{28.53} & \textbf{75.70} \\ 
\bottomrule \bottomrule
    \end{tabular}
\end{table*}

\begin{table*}[!t]
    \centering
    \caption{Benchmark of Qwen2 against the state-of-the-art S-MAD for zero-shot learning. Performance is reported in percentages, with the best results highlighted in bold.}
    \label{tab:benchmarkSMAD_SOTA}
    \begin{adjustbox}{width=\linewidth}
    \begin{tabular}{r| c c c| c c c| c c c| c c c}
    \toprule \toprule
        \multirow{3}{*}{\textbf{Approaches}} & \multicolumn{6}{c|}{\textbf{Cropped Faces}} & \multicolumn{6}{c}{\textbf{Uncropped Faces}}  \\ 
                                & \multicolumn{3}{c|}{\textbf{FERET}} & \multicolumn{3}{c|}{\textbf{FRGC}}   & \multicolumn{3}{c}{\textbf{FERET}} & \multicolumn{3}{c}{\textbf{FRGC}}  \\ 
                                &   D-EER$\downarrow$ & BSCER10$\downarrow$ & AUC$\uparrow$ &  D-EER$\downarrow$ & BSCER10$\downarrow$ & AUC$\uparrow$ & D-EER$\downarrow$ & BSCER10$\downarrow$ & AUC$\uparrow$ &   D-EER$\downarrow$ & BSCER10$\downarrow$ & AUC$\uparrow$    \\
    \midrule \midrule
MADation[TI] (Vit-B)~\cite{Caldeira-MADtion-WACV-2025} & 49.73   & 90.42 &  49.74  &  \textbf{38.78}  & \textbf{78.39} & \textbf{66.12}  &  44.26  &  83.55 & 57.09 & 36.79  & 73.88 & 68.83   \\
MADation[TI] (Vit-L)~\cite{Caldeira-MADtion-WACV-2025} & 50.47   &   90.62  &  48.81  &  51.55  & 90.71 & 47.20 &  37.03 & \textbf{72.40}  & 68.63 & \textbf{34.55} & \textbf{70.33} &  \textbf{71.90}  \\
\midrule
ours (Qwen2)                                          & \textbf{44.10}    & \textbf{88.59}  & \textbf{55.90}  &  48.96  & 89.79 & 51.04   &  \textbf{29.11}  & 81.86 & \textbf{70.89} & 39.27  & 87.26 & 60.73   \\ 
\bottomrule \bottomrule
    \end{tabular}
    \end{adjustbox}
\end{table*}

Tab.~\ref{tab:benchmarkSOTA} and Tab.~\ref{tab:benchmarkSMAD_SOTA} report a benchmark of the results of the proposed framework with those of state-of-the-art PAD and S-MAD, respectively. The baseline approaches used for the comparison were proposed in~\cite{Ozgur-FoundPAD-ArXiv-2025} (i.e., FoundPAD for PAD) and \cite{Caldeira-MADtion-WACV-2025} (i.e., MADation for S-MAD), which are based on the CLIP foundation model~\cite{Radford-CLIP-ICML-2021}.  The CLIP model has shown remarkable performance in zero-shot learning scenarios in several subsequent tasks, such as food classification, car model classification and identification of offensive memes~\cite{Radford-CLIP-ICML-2021}. These tasks involve the simultaneous use of text and image encoders for classification. For a fair comparison, the Text-Image (TI) approach proposed in both articles~\cite{Ozgur-FoundPAD-ArXiv-2025,Caldeira-MADtion-WACV-2025} is used in our work.

Note that the proposed framework significantly outperforms the state of the art by a wide margin for PAD. Our framework reports on average a HTER of 28.53\%, which is roughly half of the HTER yielded by FoundPAD (i.e., 43.97\%). A similar trend can also be observed for the AUC (75.70\% vs. 58.36\%). Notice also that OULU achieves the poorest performance among different databases, which is in line with the results shown in Fig.~\ref{fig:CD}.

Regarding the S-MAD benchmark (Tab.\ref{tab:benchmarkSMAD_SOTA}), we observe that both our framework and MADation\cite{Caldeira-MADtion-WACV-2025} perform relatively poorly when face images are cropped. In contrast, a significant performance improvement is achieved when uncropped faces are provided to the VLMs: D-EER values decrease from 44.10\% and 38.78\% to 29.11\% and 34.55\% for FERET and FRGC, respectively. These results suggest that VLMs benefit from background information when detecting morphing attacks, possibly because their pretraining involved landmark-based morphed images that often contain visible artefacts (e.g., overlapping shadows) in surrounding regions. Notably, our proposed framework using Qwen2 demonstrates competitive performance, achieving the best results in four out of twelve benchmark settings. It outperforms all baselines (i.e., MADation) on the FERET dataset in both cropped and uncropped conditions, with a D-EER of 29.11\% on uncropped FERET—the lowest D-EER among all experiments. These findings underscore Qwen2’s robustness in unconstrained scenarios, which is particularly valuable for real-world applications. While MADation~\cite{Caldeira-MADtion-WACV-2025} shows dataset-specific strengths—ViT-B-16 performing better on cropped FRGC and ViT-L-14 excelling on uncropped FRGC—Qwen2 offers more consistent performance across diverse settings. This consistency, combined with its lightweight, open-source nature, makes Qwen2 a promising candidate for scalable, zero-shot S-MAD deployment.    
  
\section{Conclusion}
\label{sec:conclusions}

In this paper, we proposed an in-context learning framework for physical (attack presentation) and digital (morphing) attack detection. The framework leverages a demonstration set that includes up to 9 different samples per category to improve the generalisability of PAD and S-MAD. By asking “Yes” or “No” questions to VLMs, the proposed approach allows computing a likelihood (similar to traditional supervised learning approaches) that avoids hallucinations and enables a systematic evaluation of these models for joint threat detection of presentation and morphing attacks. 

The experimental evaluation was conducted in compliance with the metrics defined in the ISO/IEC 30107-3~\cite{ISO-IEC-30107-3-PAD-metrics-2023} and ISO/IEC 20059~\cite{ISO-IEC-20059} on well-established and commonly used databases and protocols for PAD and S-MAD. Three different publicly available VLMs (i.e. Qwen2-VL, Idefics3 and MiniCPM) were evaluated for both types of attacks, leading to different findings:

\begin{itemize}
    \item Qwen2 reported, among the VLMs, the best generalisation capability in the detection of unknown physical and digital attacks using only a few samples during inference: D-EERs for PAD are in the range 10\%-26\% for unknown PAI species, and the mean HTER was 28.53\% for the cross-database scenarios. Down to 9\% of D-EER was reported for S-MAD without any demonstration set (i.e., zero-shot inference).
    
    \item It was demonstrated that the background context significantly improved S-MAD performance: performance improvements went from 49\% to 9\% in terms of {D-EER} depending on the morphing tools when using an uncropped face.   

    \item A benchmark of the proposed framework against the state-of-the-art in both zero-shot PAD and S-MAD demonstrated a significant performance improvement. For PAD, our approach achieved an average HTER of 28.53\%, substantially outperforming the current state-of-the-art FoundPAD~\cite{Ozgur-FoundPAD-ArXiv-2025}, which reported an HTER of 43.97\%. In the case of S-MAD, while MADation~\cite{Caldeira-MADtion-WACV-2025} exhibited dataset-specific strengths—particularly with ViT-B-16 performing better on cropped FRGC and ViT-L-14 on uncropped FRGC—our Qwen2-based framework outperformed all baselines on the FERET dataset in both cropped and uncropped settings. Notably, it achieved the lowest D-EER (29.11\%) across all benchmarked experiments on uncropped FERET, underscoring its robustness in unconstrained environments.
    
    \item In general, both algorithms (i.e., MADation and our Qwen2-based framework) show a notable performance improvement when facial images are provided uncropped. Specifically, the average D-EER decreased from 44.57\% (cropped) to 34.18\% (uncropped), while the average AUC increased from 56.38\% to 64.84\%. This corresponds to an improvement of up to 10.4 percentage points in AUC and a reduction of over 10\% in D-EER, highlighting the positive impact of background information on zero-shot S-MAD performance. 
\end{itemize}

As the framework is flexible, it can be combined with any VLM. Therefore, we expect a significant improvement in PAD and S-MAD performance if combined with other large VLMs such as GPT-4o and Gemini 2.0, which have extensive knowledge of PAD and MAD concepts (\cite{Komaty-ChatGPTPAD-ArXiv-2025,Zhang-ChatGPTSMAD-2025}). In future work, we plan to extend this framework to D-MAD, which uses image pairs to detect the attack.

\section*{Acknowledgment}

This research work has been partially funded by the European Union (EU) under G.A. no. 101121280 (EINSTEIN) and CarMen (101168325), and the German Federal Ministry of Education and Research and the Hessian Ministry of Higher Education, Research, Science and the Arts within their joint support of the National Research Center for Applied Cybersecurity ATHENE.

\ifCLASSOPTIONcaptionsoff
  \newpage
\fi



\bibliographystyle{IEEEtran}
\bibliography{IEEEfull}

\vfill

\end{document}